\documentclass[10pt,journal,compsoc]{IEEEtran}

\ifCLASSOPTIONcompsoc
  \usepackage[nocompress]{cite}
\else
  \usepackage{cite}
\fi

\ifCLASSINFOpdf
\else
\fi

\usepackage{graphicx}
\usepackage{amsmath}
\usepackage{url}

\begin{document}

\title{Applying Probabilistic Programming \protect \\ to Affective Computing}

\author{Desmond~C.~Ong,~\IEEEmembership{Member,~IEEE~Computer~Society,} 
        Harold~Soh,~\IEEEmembership{Member,~IEEE~Computer~Society,} \protect\\
        Jamil~Zaki, and~Noah~D.~Goodman%
\IEEEcompsocitemizethanks{
\IEEEcompsocthanksitem D. C. Ong is with the A*STAR Artificial Intelligence Initiative and with the Institute of High Performance Computing, Agency of Science, Technology and Research (A*STAR), Singapore 138632. 
E-mail: desmond.c.ong@gmail.com
\IEEEcompsocthanksitem H. Soh is with the Department of Computer Science, National University of Singapore, Singapore 117417.
\IEEEcompsocthanksitem J. Zaki is with the Department of Psychology, Stanford University, Stanford CA 94305.
\IEEEcompsocthanksitem N. D. Goodman is with the Department of Psychology and the Department of Computer Science, Stanford University, Stanford CA 94305.}%
}

\markboth{IEEE Transactions on Affective Computing, Manuscript ID}%
{Ong \MakeLowercase{\textit{et al.}}: Applying Probabilistic Programming to Affective Computing}


\IEEEtitleabstractindextext{%
\begin{abstract}
Affective Computing is a rapidly growing field spurred by advancements in artificial intelligence, but often, held back by the inability to translate psychological theories of emotion into tractable computational models. To address this, we propose a probabilistic programming approach to affective computing, which models psychological-grounded theories as generative models of emotion, and implements them as stochastic, executable computer programs. We first review probabilistic approaches that integrate reasoning about emotions with reasoning about other latent mental states (e.g., beliefs, desires) in context. Recently-developed probabilistic programming languages offer several key desidarata over previous approaches, such as: (i) flexibility in representing emotions and emotional processes; (ii) modularity and compositionality; (iii) integration with deep learning libraries that facilitate efficient inference and learning from large, naturalistic data; and (iv) ease of adoption. Furthermore, using a probabilistic programming framework allows a standardized platform for theory-building and experimentation: Competing theories (e.g., of appraisal or other emotional processes) can be easily compared via modular substitution of code followed by model comparison. To jumpstart adoption, we illustrate our points with executable code that researchers can easily modify for their own models. We end with a discussion of applications and future directions of the probabilistic programming approach
\end{abstract}

}

\maketitle

\IEEEdisplaynontitleabstractindextext

\IEEEpeerreviewmaketitle

\IEEEraisesectionheading{\section{Introduction}\label{sec:introduction}}

\IEEEPARstart{A}{CHIEVING} a human-like understanding of emotions is a holy grail of affective computing. The ideal affective computer has to correctly identity a user's emotion based on behavioral cues and contextual information, reason about how its actions may affect the user's emotions, and choose its responses accordingly. Affective agents deployed in a variety of applications, such as in tutoring \cite{dmello2007toward}, social robotics \cite{vallverdu2009handbook}, or other human-computer interactions \cite{brave2003emotion}, are aiming to achieve this next frontier of emotional understanding. Research in recent years has made important strides towards this goal, producing hundreds of papers on applying machine learning techniques to recognizing emotions (see \cite{calvo2010affect, gunes2010automatic, jaimes2007multimodal, zeng2009survey} for reviews). The success of these efforts is due in large part to recent developments in deep learning algorithms and computational power \cite{kim2013deep}, coupled with the availability of larger datasets \cite{douglas2007humaine}. However, though these machine learning models may demonstrate excellent performance at emotion recognition, they nevertheless fall short of being able to do true reasoning \cite{ong2018computational}. They are unable to carry out counterfactual and hypothetical reasoning, provide causal attributions and explanations, or incorporate contextual knowledge into their inferences. And they usually do not generalize well outside the dataset they are trained on. By merely performing pattern recognition, these machine learning models achieve \emph{perception}, but fall short of \emph{cognition} about affect.

In contrast to the data-driven, machine learning approach, an alternative approach with a much older history in affective computing favors constructing theory-based models, such as emotion architectures (see \cite{lin2012computational, marsella2010computational} for reviews). These models are inspired by psychological theories of emotion (e.g., \cite{frijda1986emotions, smith1993appraisal}), and focus in detail on modeling the components and the computations that go into emotional processes. For example, many such models focus on computationally defining the cognitive evaluations of experienced situations---\emph{appraisals}---that give rise to emotions \cite{broekens2008formal, marsella2009ema, ortony1990cognitive}. Other models also focus on modelling how emotions influence cognition and subsequent behavior \cite{becker2008affect, marinier2009computational}. However, these models tend to be hand-tuned to specific theories and specific contexts---for example, the model in \cite{ortony1990cognitive} contains many rule-based appraisals that give rise to emotions, but is not able to learn new appraisals or modify existing appraisal rules for new contexts---and thus they are unable to scale well to the complexity of larger, naturalistic datasets. Another limitation is that these models usually do not specify how one goes from naturalistic data, such as a pixel-level representation of a smiling face, to a representation of emotion: The exact ``transformation" from an emotion to a visually-observable muscular configuration may be scientifically uninteresting (except perhaps for the fact that such a reliable and valid mapping does or does not exist), yet still important to an engineer wishing to build applications that are sensitive to affect.

These theory-driven models differ not only in their theoretical assumptions and content, but also in the details of their implementation, such that it is problematic to compare theories side-by-side \cite{lin2012computational, marsella2010computational}. To add to this difficulty, theory-driven approaches tend to have large barriers to adoption, as most of them use specific architectures or systems that may be difficult for other researchers to adopt. For example, the EMA model \cite{marsella2009ema} and Marinier and colleagues' \cite{marsella2009ema} model are built on top of SOAR \cite{newell1992soar}, and a researcher who wishes to contribute to these lines of work will have to learn an unfamiliar set of syntax and conventions---although some have advocated for the field settling on a unified architecture \cite{reisenzein2013computational}. Other research groups may implement their models in custom-built systems \cite{becker2008affect, dias2014fatima}, which may not be easily available for other researchers to build upon.

By contrast, deep learning and other machine learning approaches have experienced a meteoric rise in popularity and adoption over the past few years. This is due in large part to readily-available, open-source, deep learning tools written in beginner-friendly programming languages like Python and R, along with accessible learning materials (tutorials, online courses) and a large and active community (on programming forums and blogs like Stack Overflow and Medium respectively). An ideal approach is one that ``democratizes" affective computing by being easier to adopt, which will increase the speed of research as well as infrastructural developments, while still maintaining the scientific rigor and experimentation of theory-based approaches.

In this paper, we propose a probabilistic programming approach to affective computing that marries the strengths of theory-driven approaches with that of data-driven approach. Probabilistic programming is a modelling paradigm by which one can specify theories of emotion using probabilistic, generative models \cite{goodman2013principles, goodman2008church}. We can explicitly represent uncertainty in emotion theory, randomness in emotional phenomena, or even incomplete knowledge about an agent's mental state, as programs that contain some degree of randomness. Because probabilistic programs are modular and can be composed to form more complex programs, we can focus on modeling at different levels of abstraction \cite{marsella2010computational, reisenzein2013computational}. Having a hierarchy of abstract representations also allows probabilistic programs to learn context-specific knowledge (i.e., to the specific example) as well as knowledge that can generalize to other scenarios \cite{tenenbaum2011grow}. One can use a probabilistic programming framework to test different emotion theories (e.g., of appraisal) by substituting modular chunks of code and testing which theories best fit experimental data. Modularity also allows integration of emotion with other high-level theories of psychological phenomena, such as mental states and motivation \cite{ong2018computational, saxe2017formalizing}.

In probabilistic programming, model specification is orthogonal to learning and inference in these models. This separation allows the affective computing modeler to focus on specifying the model rather than on inference and optimization methods, much like how so-called ``high-level" programming languages abstract away the workings of machine code from the programmer. Such abstraction lowers the barrier to adoption by making languages easier to learn, while simultaneously making them more efficient\footnote{For example, object-oriented programming was a major abstraction that allowed complex data structures (objects) with their own functions.}. Because inference in probabilistic programming is orthogonal to model specification, infrastructural development of inference algorithms can proceed in parallel \cite{ritchie2016deep}. Indeed, many modern probabilistic programming languages leverage existing deep learning, optimization, and inference libraries, which allows efficient and scalable learning from large datasets \cite{tran2017deep}. The modeller can simply specify the model, give it data, and press ``run", so to speak, relying on general-purpose inference implemented within the probabilistic programming language. Our claim is that probabilistic programming combines the strengths of the data-driven and theory-driven approaches in affective computing: It allows the building of psychologically-grounded models, hypothesis testing and scientific experimentation, within an infrastructure to learn efficiently from and do inference over large data. And it provides a common platform with a low barrier to entry that will encourage integration among existing approaches in the field.

We begin the paper by introducing the \emph{intuitive theory} approach to understanding reasoning about emotion, and how this approach has been formalized in the computational cognitive science literature using probabilistic methods---specifically probabilistic graphical models. Next, we introduce probabilistic programming by discussing two recent high-impact examples of applying probabilistic programming to model human cognition. With this background, we then discuss implementing a computational model of emotion using probabilistic programming, by providing worked example code in a re-analysis of a small multimodal dataset \cite{ong2015affective}. We illustrate how one can model components such as appraisal and emotion recognition from faces. We also demonstrate reasoning capabilities that this probabilistic programming approach has over previous approaches: for example, it can generate novel emotional faces given a new situation. Finally, we end by discussing the boundaries of this approach, as well as the long-term promise of this approach, such as to modelling emotion generation and understanding complex intentional emotional displays.

\section{Implementing Intuitive Theories as \\ Probabilistic Models}

People have an intuitive understanding of the world around them. The average person may not be able to write down Newton's laws, but, upon witnessing a thrown baseball, is able to intuitively predict the ball's trajectory and where it would land \cite{gerstenberg2017eye, mccloskey1983intuitive}. This intuitive understanding also extends to making sense of other people. If we see a roommate walk out of their room, pause, and turn back, we intuitively start generating possible hypotheses about their behavior---perhaps they forgot something in their room and went back to retrieve it---even though these inferences are made on sparse and incomplete information. In daily life, we effortlessly make such inferences about other people's thoughts, feelings, and even personality and other traits \cite{gilbert1998ordinary, heider1958psychology, ross1977intuitive}. Such ``intuitive physics" and ``intuitive psychology" are made possible by intuitive theories that consist of a structured ontology of concepts---\emph{gravity}, \emph{air resistance}, and \emph{velocity}; \emph{goals}, \emph{personality}, and \emph{behavior}---and the causal relationships between these concepts \cite{gerstenberg2017intuitive, gopnik1997words}. This knowledge allows us to make predictions, complex inferences, and offer explanations about the observed world. If we know that Bob wants to eat (\emph{goal}), we can predict that he will go look for food (\emph{behavior}). Conversely, if we see that Bob has gone to the cupboard and returns with some potato chips, we can also infer that he knew that there were snacks in the cupboard (Bob's \emph{beliefs} about the world), and perhaps that he likes potato chips (\emph{preferences}) \cite{baker2017rational}. Much like how scientific theories allow scientists to produce a coherent description of natural phenomena, these intuitive theories allow humans to reason about and explain both the physical and the social world that they live in \cite{gopnik1997words, wellman1992cognitive}.

In recent work, we \cite{ong2015affective, ong2018computational} and others \cite{saxe2017formalizing, wu2018rational} have proposed that people also possess a structured intuitive theory of emotions. This intuitive theory comprises conceptual knowledge such as: what are emotions, moods and other affective states; what types of event outcomes cause emotions; what mental states influence emotions; and what types of behavior emotions influence. Importantly, this conceptual knowledge goes beyond encoding simple pairwise emotion-behavior contingencies, such as recognizing emotional facial expressions, which many machine learning models are already adept at \cite{zeng2009survey}. It allows reasoning about emotions in more complicated situations, such as with novel events or in a new (e.g. cultural) context, and enables explanation generation, counterfactual reasoning, and hypothetical reasoning \cite{ong2018computational}. Laypeople use their rich intuitive theories effortlessly in daily life---the challenge for researchers is to distill this implicit knowledge into workable computational models.

We note that there are important differences between scientific theories of emotion (e.g. \cite{ellsworth2003appraisal, ortony1990cognitive}) and the intuitive theories that laypeople employ. For one, the lay description of emotion is far more coarse-grained. Intuitive theories do not care about the neural bases of emotion \cite{lindquist2012brain}, nor are they concerned with a taxonomy of how many emotions there are, or whether there is a hierarchy between ``basic" and other emotions \cite{ekman1992argument, ortony1990whats}. In fact, these intuitive theories differ remarkably across cultures---the antecedents and consequents of emotions like shame vary by culture \cite{wierzbicka1999emotions, wong2007cultural}---as well as across individuals \cite{ong2018happier}, based on 
mood \cite{devlin2014not} or mental illness \cite{devlin2016tracking}. In other words, while scientific theories seek an ``objective" truth, lay theories describe the subjective reality that people live in. 
Thus, it is critical to model intuitive theories because they are what laypeople use to make sense of others' emotions.

The intuitive theory approach lends itself well to computational modeling, especially probabilistic graphical modeling. A probabilistic graphical model represents random variables as well as the probabilistic interdependencies between these variables \cite{koller2009probabilistic}. For example, a widely supported intuitive theory of human behavior holds that (people think that other) people have \emph{beliefs} about the state of the world, have \emph{desires} (i.e., goals that they want to achieve), and subsequently form \emph{intentions} to act upon their beliefs to maximize their desires \cite{dennett1989intentional, malle2006mind}. This belief-desire psychology, or belief-desire-intention psychology, can be formalized as a rational agent acting to maximize its utility given its incomplete knowledge about the world, such as using decision networks or partially-observable Markov decision processes \cite{baker2009action, jara2016naive, jern2015decision}. Inferring beliefs and desires from actions reduces to Bayesian inference over this generative model \cite{baker2017rational}. These recent ``Bayesian Theory of Mind" models have been successful at predicting participants' judgments of agents' beliefs and desires, suggesting that they capture how people reason about goal-directed behavior.

We can also add emotions into these generative models of behavior. People know that others' emotions arise as a reaction to a motivationally salient event in the world, based upon their beliefs and desires about the world \cite{scherer2008facial, van2010interpersonal}. Many scientific theories of emotion propose that people implicitly evaluate (``\emph{appraise}") experienced events along a number of self-relevant dimensions based upon their mental state, and feel emotions as a consequence of this appraisal \cite{ellsworth2003appraisal, ortony1990cognitive, smith1993appraisal}. For example, a goal-conducive event like winning a prize would produce happiness, while its unexpectedness could also give rise to surprise. Recent work has also suggested that laypeople perform a similar, ``third-person appraisal" process to reason about the emotions of others \cite{demelo2014reading, ong2015affective, skerry2015neural, van2010interpersonal, wondra2015appraisal, wu2018rational}. When reasoning about someone else's emotions, people implicitly adopt that individual's perspective and appraise the situation on behalf of that individual. This capability develops very early in life: even infants and young children are able to reason about the emotional expressions that accompany fulfilled or thwarted goals \cite{skerry2014preverbal, wellman2000young} or mismatched expectations \cite{ong2016young, wu2018inferring}. Thus, a simple intuitive theory of emotions might have event outcomes and mental states jointly ``causing" emotions via an appraisal process, which in turn ``cause" facial expressions and other behavior \cite{demelo2014reading, ong2015affective, saxe2017formalizing, wu2018rational}.

\begin{figure}[!bt]
\centering
\includegraphics[width=.8\columnwidth]{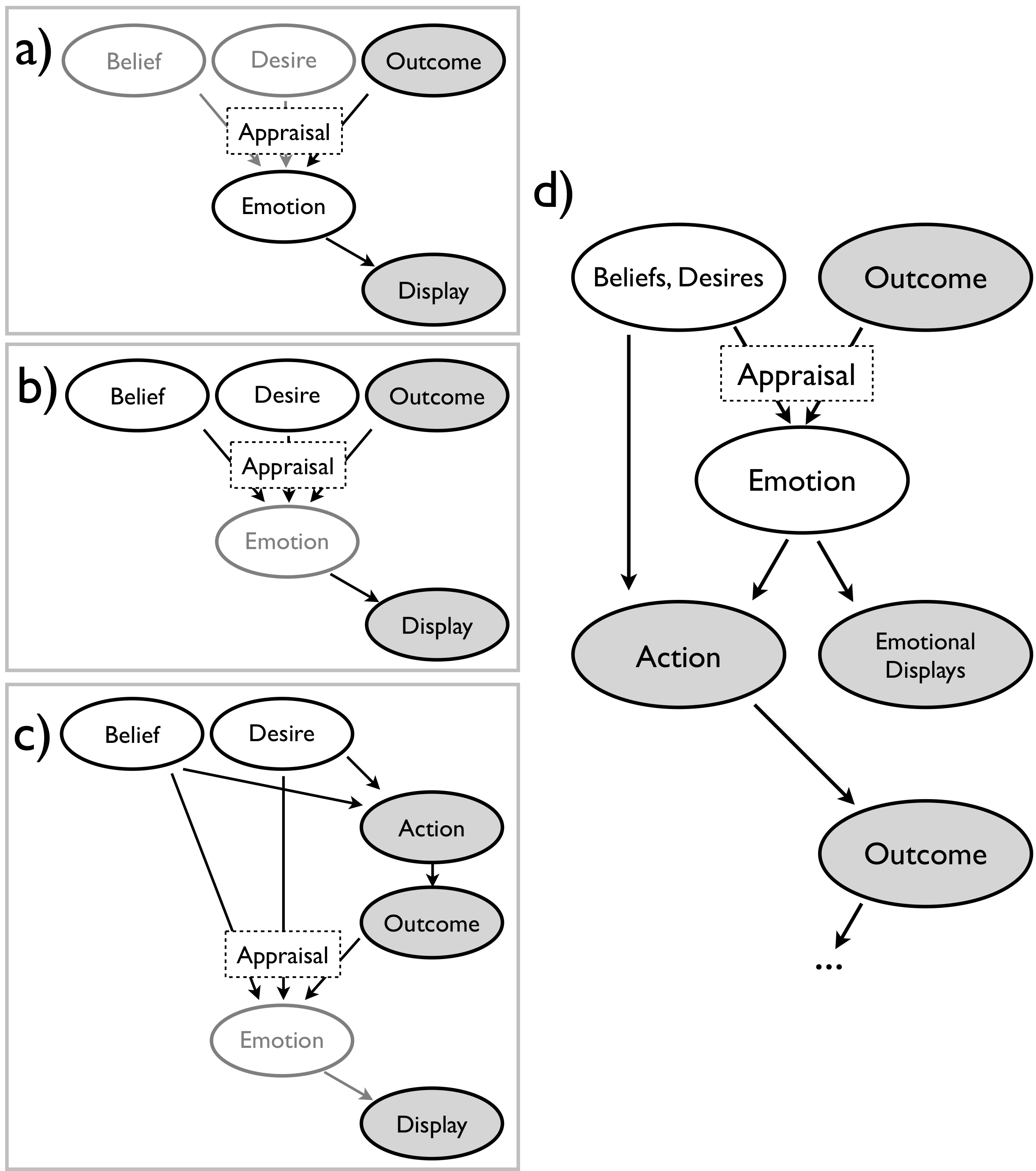}
\caption{Summary of previous work using probabilistic graphical approaches to model lay human reasoning about emotions. Figures are adapted from (a) \cite{ong2015affective}, (b) \cite{demelo2014reading}, (c) \cite{wu2018rational}. Shaded nodes represent observable variables, while unshaded nodes represent latent variables. Variables that were ``implicitly" modelled (i.e., not directly specified) are rendered using translucent nodes. In (d), adapted from \cite{ong2018computational} and \cite{saxe2017formalizing}, we illustrate the more general model that encompasses the models in (a-c).}
\label{fig:LTmodels}
\end{figure}

Several recent studies modeled such intuitive reasoning about emotions using probabilistic graphical models (Fig. \ref{fig:LTmodels}). We \cite{ong2015affective} investigated the third-person appraisal process as people evaluated the emotions of others playing gambles. We also had participants reason about emotions given both the gamble outcome and another behavioral cue (e.g., facial expression or verbal utterance). Such complex inferences from multiple cues---\emph{cue integration}---can be mathematically derived using Bayesian inference \cite{zaki2013cue}, and the predictions from a Bayesian model track laypeople's judgments in these complex scenarios. De Melo and colleagues \cite{demelo2014reading} studied laypeople's inferences of appraisals from emotional expressions by modeling the generative processes from outcomes to appraisals, emotions, and expressions. Given an emotional expression, one can compute an estimate of the latent appraisals by doing reverse inference in the model, a process they term ``reverse appraisal". Indeed, they find that laypeople's subsequent behavior towards the agent are mediated by their inferred appraisals of the agent's emotions. Wu and colleagues \cite{wu2018rational} similarly proposed a model of how mental states---beliefs and desires---give rise to actions and emotions; They model inference of mental states from emotional expressions via Bayesian inference, and show that these also track participants' judgments. Although these independently-conducted studies contain differences in the details of their models (Fig. 1), there was surprising agreement in the broader approach of using Bayesian inference on a generative model of emotion, and even in the general causal flow of the model. For example, each study highlighted the importance of the third-person appraisal process as an antecedent to emotion (for a more in-depth discussion, we invite readers to see \cite{ong2018computational}).

From a computational cognitive science standpoint, inference in these probabilistic models of emotions track laypeople's judgments, showing that such models provide a good computational account of laypeople's psychology, yielding valuable scientific insight. From an affective computing standpoint, such approaches afford strong, theory-grounded models upon which to build applications that reason like laypeople, and that laypeople reason about. For example, if the goal is to build an expressive virtual character \cite{swartout2006toward} or affective tutor \cite{dmello2007toward} that interacts with laypeople, then perhaps an intuitive theory-based model may provide a way to generate more human-like affective behavior.

Implementing intuitive theories as probabilistic graphical models has its limitations. In our opinion, the most significant limitation is \emph{representation}. Variables like emotion are often represented using a single real-valued number, or a vector along a number of emotion dimensions (see also \cite{yannakakis2017ordinal}). These representations cannot easily encode complex information like the relational and temporal nature of emotions: John is not just angry, 6 out of 7, he is often angry \emph{at} something or someone; This anger may be fleeting, or may give rise to a life-long resentment. Such information is crucial for proper reasoning about what behavior John might next exhibit or how one might intervene to help regulate John's emotions, but is difficult to represent using graphical models. Other representational challenges arise when considering other affective phenomena, such as moods and appraisal. Consider appraisal: Within a constrained scenario such as playing a gamble or a social dilemma game with another agent, the appraisal process may reduce to some linear combination of real-valued features that define the outcome of the game. More generally, however, the appraisal process is a complex evaluative process that is difficult to represent with current graphical approaches. Finally, learning probabilistic graphical models from data and doing inference in large models is computationally intensive, limiting their real-world applications on larger, naturalistic datasets. We propose that a modern ``successor" of probabilistic graphical modeling---probabilistic programming---may offer a solution to these limitations.

\section{Implementing Intuitive Theories using Probabilistic Programming}

\begin{figure}[!tb]
\centering
\includegraphics[width=.9\columnwidth]{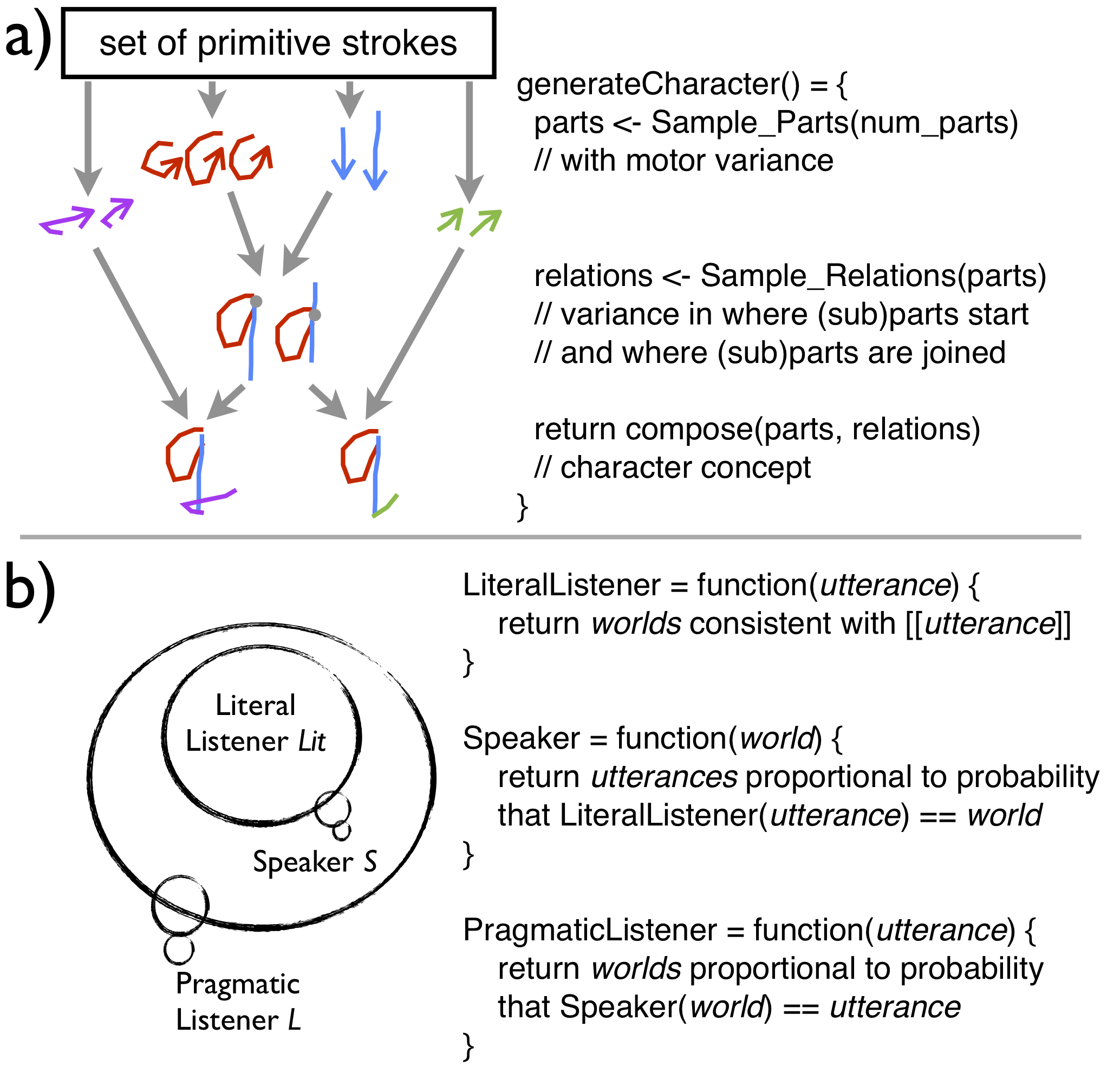}
\caption{Simplified illustration of probabilistic programming applied to human cognition, with pseudocode. (a) Our adaptation of the key ideas in \cite{lake2015human}. Characters (e.g., ``g" and ``q") are composed of a sequence of parts and subparts and their spatial relations (i.e., where the parts are joined). There is motor variance in the ``writing" process, which may lead to visually similar, but conceptually different, character concepts. (b) Language understanding in the Rational Speech Acts framework \cite{goodman2016pragmatic} proceeds via nested reasoning: the pragmatic listener reasons about a nested speaker, which in turn reasons about a nested literal listener.}
\label{fig:ppExamples}
\end{figure}

Probabilistic programming is a relatively new and powerful modeling paradigm that offers much promise for affective computing. Like probabilistic graphical modeling, probabilistic programming allows one to capture abstract, conceptual knowledge (e.g., in human intuitive theories) as generative models. Instead of a graphical representation, probabilistic programming represents conceptual knowledge as stochastic programs---chunks of code that embed randomness into their execution \cite{goodman2008church, goodman2014concepts}. We start by introducing two recent examples of probabilistic programming applied to model human cognition (Fig. \ref{fig:ppExamples}), before discussing its features and its application to affective computing.

In a recent landmark study, Lake, Salakhutdinov, and Tenenbaum \cite{lake2015human} introduced a handwriting-recognition model that successfully learns character concepts. After being shown just one example of a novel, handwritten character, the model was able to correctly identify more examples at a level comparable to humans (one-shot classification); and when made to generate new exemplars of a novel character, produced examples that were indistinguishable from those produced by human volunteers (human-like generative capacity). Underlying these impressive capabilities is a powerful idea: a probabilistic generative model that models the actual writing process. The model has, as primitives, handwritten strokes such as straight lines or curves. These strokes are composed to form more complex parts that in turn are composed to make characters, subject to constraints on where and how different strokes may be joined together to form different parts or characters. Finally, the model allows for motor variance at different steps in the handwriting process, such as in choosing the start point of each stroke, trajectory of each stroke, and how different strokes are joined together. Such variance in the generation process introduces ``noise" into the visual appearance of the character, but do not change the underlying concept that is to be conveyed---a badly-written ``g" is still a ``g", despite its potential visual similarity to ``q" (Fig. \ref{fig:ppExamples}a). When presented with novel characters, the model infers the steps needed to generate the input characters: it inductively learns a \emph{program} for producing the character. This allows it to learn abstract conceptual knowledge from visual features, and to flexibly apply this knowledge in a more human-like manner \cite{lake2017building}.

Probabilistic programming can also be applied to social reasoning, such as language understanding \cite{goodman2013knowledge}. In conversation, people naturally assume, by convention, that others are being informative when speaking, and this allows people to make pragmatic inferences over what was not explicitly said \cite{grice1975logic}. A statement like ``Some of the students passed the exam" invites the interpretation ``Not all of the students passed the exam", even though the semantics of the latter statement is stricter than and not necessarily implied by the former \cite{potts2016embedded}. We and our colleagues have formalized this reasoning in terms of a Rational Speech Act (RSA) framework for language understanding \cite{goodman2016pragmatic}. First, we define the literal listener (\emph{Lit}) as a probabilistic program that maps a heard utterance to its possible literal meanings (e.g., ``some" denotes a non-zero quantity, which could be \emph{one}, or \emph{two}, etc.). Next, the speaker (\emph{S}) is a probabilistic program that models a cooperative human communicator: It aims to achieve the goal of \emph{Lit} correctly inferring the state of the world, and chooses an utterance to achieve that goal. Finally, the pragmatic listener (\emph{L}) hears an utterance and reasons about the goals of the speaker \emph{S} who produced it. Thus, we implement the pragmatic listener \emph{L} as a probabilistic program that has within it a nested program \emph{S}, which has in turn another nested program \emph{Lit} (Fig. \ref{fig:ppExamples}b). This allows nested social reasoning: the listener reasoning about the speaker reasoning about the listener. For example, upon hearing ``Some of the students passed the exam", \emph{L} reasons that, if it were indeed the case that ``all the students passed the exam", then \emph{S} would have said so; since \emph{S} could have but did not, it is more likely that ``Some, but not all of the students" achieved a passing grade. Note that \emph{S} and \emph{Lit} are not actual agents, but they exist within listener \emph{L}'s intuitive theory of communication, which makes assumptions that speakers choose their behavior rationally and following Gricean maxims. The RSA framework has been applied to model understanding of generic language \cite{tessler2016pragmatic}; nonliteral language like hyperbole \cite{kao2014nonliteral}; humor in wordplay \cite{kao2016computational}; and politeness in indirect speech \cite{yoon2017won}. More generally, using probabilistic programs in a compositional---in this case, nested---manner provides a computational framework for modeling social reasoning.

The two examples above illustrate some of the features of probabilistic programming as a modeling paradigm that make it appealing to affective computing. The core idea is representing theory---laypeople's intuitive theories about handwriting, communication, or \emph{emotions}---in terms of probabilistic programs. Unlike deterministic programs that always produce the same output when given the same input, probabilistic programs instead produce samples from a distribution of possible outputs. This allows explicit modeling of uncertainty, whether such uncertainty arises from (i) incomplete knowledge about the world and others' unobservable mental states, (ii) incomplete theory, or (iii) inherent randomness in the generative process. For example, Lake and colleagues \cite{lake2015human} introduce one form of uncertainty via motor variance at various steps in the character generation process, while in the RSA framework \cite{goodman2016pragmatic}, the model explicitly represents uncertainty in semantic meaning and speaker goals. Affective computing applications face many sources of uncertainty. Third-person appraisal requires an uncertain inference about others' latent beliefs and desires (incomplete knowledge). There are individual differences, such as in personality or cultural background, in how people with the same expectations and goals appraise the same outcomes, and how they will behave after: Many of these individual differences have not been explored by scientists (incomplete theory). Finally, the same person facing the same situation in the same context might not always behave in the same way (inherent randomness). Explicitly modelling these sources of uncertainty is important for learning from data and generalizing to new agents and contexts.

A second crucial feature is \emph{modularity}. In a probabilistic programming paradigm, small, modular probabilistic programs are composed together in a hierarchical and/or sequential fashion to produce more complex phenomena---or to capture reasoning about such phenomena. In the handwriting recognition example, there is a hierarchy where strokes compose to form sub-parts, parts, and characters, while in the language understanding example, social reasoning proceeds via inference in nested programs. Modularity and compositionality are particularly important in modeling emotions and the processes that give rise to and arise from emotion \cite{marsella2010computational, reisenzein2013computational}. This is because modeling emotion necessitates modeling a wide range of processes from emotion elicitation (via appraisal) to the behavioral effects of emotion. Modularity is essential for defining small, re-usable processes, and compositionality allows one to structure complex reasoning over these processes. For example, an agent experiencing a negative emotion (after appraisal) may want to down-regulate their negative emotion. One way that it can do that is to choose an action that will likely result in positive appraisals; Thus, the agent has to re-use its concept of appraisal to best select an action to achieve its goals. In fact, this action selection is an example of an inference that is handled naturally in a probabilistic program. With a generative model of actions to outcomes to appraisals to emotions, one can condition on a desired emotion, infer desired appraisals and outcomes, and infer the actions that one has to take to achieve the desired goal.

Probabilistic programming is also becoming easier to implement. There has been a surge of development in probabilistic programming infrastructure in the past decade, due in part to a large funding initiative from DARPA \cite{roberts2013ppaml}. There now exists many probabilistic programming languages under active development, and many of these exist as modules or libraries written in existing programming languages like Python (e.g., Pyro \cite{bingham2018pyro} and Tensorflow Probability) and Javascript (e.g., WebPPL \cite{goodman2014design}). These languages are Turing-complete, and can represent any computable probability distribution. Moreover, some of these language leverage existing optimized deep-learning libraries: As its name suggests, Google's Tensorflow Probability is built on top of Tensorflow, while Uber-developed Pyro is built on top of PyTorch. This allows one to leverage efficient optimization (e.g., gradient descent algorithms), approximate inference techniques (e.g., MCMC, variational inference), as well as hardware acceleration (e.g., GPU computation). Thus, these modern probabilistic programming languages combine both the modeling flexibility of a universal programming language with the power of modern deep learning.

\section{Modeling Emotions using Probabilistic Programs}

At an abstract level, probabilistic programming is an approach to modeling: It is not bound to any particular language or even a particular architecture, unlike work in cognitive and emotion architectures \cite{marinier2009computational, marsella2009ema}. That said, to illustrate our points more concretely and to jumpstart the community's adoption of a probabilistic programming approach to affective computing, in this section we provide accompanying code written in the open-source probabilistic programming language Pyro, itself written in Python. A repository for the code in this paper, accompanying documentation, and links to tutorials, are available at: \\ \texttt{\url{https://github.com/desmond-ong/pplAffComp}}

\subsection{Description of Dataset}

As an illustrative example throughout this section, we use a previously-collected dataset \cite{ong2015affective} (Experiment 3; available at \texttt{\url{github.com/desmond-ong/affCog}}). 
In this experiment, participants were shown characters playing gambles for money. On some trials, participants saw the outcome of the gamble, or a facial expression ostensibly made by the character after seeing the gamble, or both the outcome and the facial expression. They then rated how the character felt on 8 emotions (e.g., \emph{happy}, \emph{sad}, \emph{anger}), each using 9 point Likert scales. The facial expressions were all generated using FaceGen, with standardized gender, race, and other features, and varying only in emotional expressions. Previously \cite{ong2015affective}, we used a probabilistic graphical modelling approach to show that participants' judgments of emotions given multiple cues---what we term emotional cue integration---can be modeled as the joint Bayesian inference of P(\emph{emotion}$|$\emph{outcome}, \emph{face}), using the individual likelihoods P(\emph{emotion}$|$\emph{outcome}) and P(\emph{face}$|$\emph{emotion}).

In this paper, we use probabilistic programming to re-model this dataset. The purpose of this re-modeling is pedagogical. We use a real dataset (rather than a simulated, toy dataset) of managable size and with actual hypotheses. Our aim is not to outperform our prior analysis, but to provide readable, illustrative examples for high-level takeaways. However, we show later that the probabilistic programming re-modeling offers more capabilities than our prior analysis, such as the ability to generate novel faces.

\subsection{Modeling Appraisal}

First, let us consider the ``causes" of emotion---the appraisal process. Appraisal can be represented by a function that takes in a representation of the event and performs some computation to yield emotions. The model could also define an intermediate representational space of appraisal dimensions, such as \emph{goal-conduciveness}, \emph{novelty}, and \emph{controllability} (e.g., \cite{broekens2008formal, ortony1990cognitive}), map events onto those dimensions, and subsequently map from the appraisal dimensions to emotions.

\begin{figure}[!tb]
\centering
\includegraphics[width=\columnwidth]{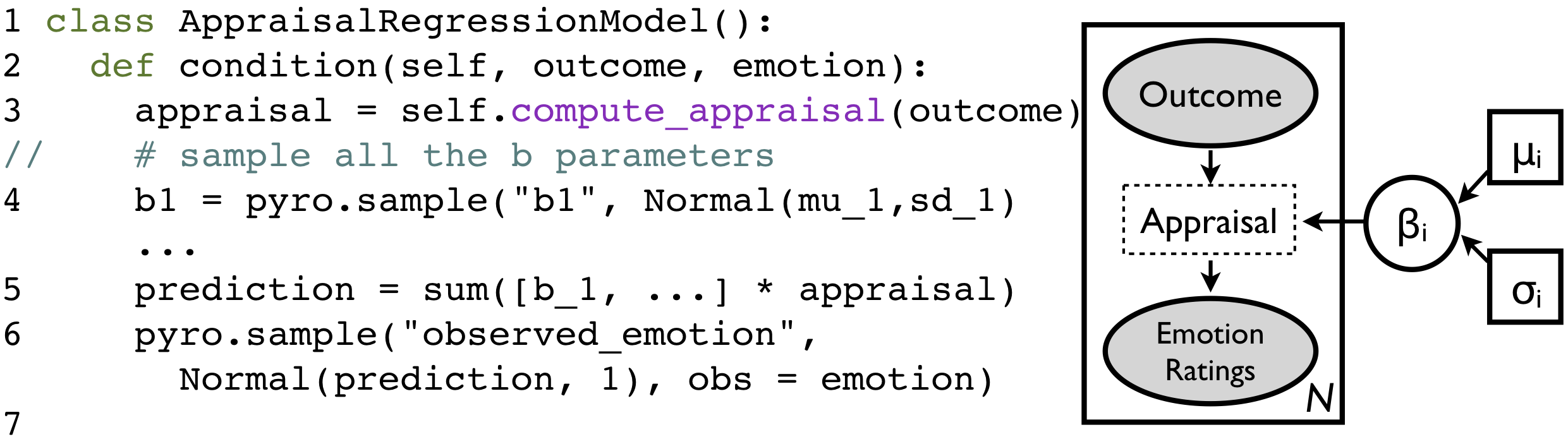}
\caption{Excerpt of Pyro code that learns a linear regression mapping appraisals to emotion ratings, with graphical representation on the right. We use plate notation: There are $N$ independent observations of outcomes and emotion ratings, and the parameters $\beta$ are constant and shared across the observations. The \texttt{compute\_appraisal()} function takes in a representation of an outcome and returns an appraisal ($l$. 3). We then sample regression coefficients $\beta_i$ for each dimension $i$ from a Normal distribution, given parameters $\mu_i, \sigma_i$ ($l$. 4). We compute the estimated emotion rating ($l$. 5), and then condition on having observed the emotion rating in the data ($l$. 6), in order to infer the values of $\mu_i, \sigma_i$.}
\label{fig:code1}
\end{figure}

Following our previous work \cite{ong2015affective}, we consider a simple linear model (Fig. \ref{fig:code1}). People observe an agent experiencing the outcome of a gamble, and provide emotion ratings of how that character is feeling. We assume that to do so, observers compute an appraisal of the outcome via a linear function of the features of the gamble. We can, based on theory, identify a set of features of the gamble that may factor into such an appraisal, such as the amount won relative to the expected value of the gamble. Let us abstract out that computation into a \texttt{compute\_appraisal()} function that returns a set of appraisal variables. Hence, we have a model that learns a linear mapping from these appraisal features to the emotion---really, a linear regression. 

 
This linear model provides a simple starting point that already contains many rich features. We abstracted away the appraisal calculation into a separate \texttt{compute\_appraisal()} function, which we can modify without impacting the logic of the generative model. Here, we modeled the mapping from appraisal variables to emotion ratings using a linear regression; we could substitute that with a non-linear function, like a feed-forward neural network that we fit to the data. This choice of fitting and learning is conceptually distinct from the choice about appraisal dimensions. 
 
\begin{figure*}[!tb]
\centering
\includegraphics[width=.8\textwidth]{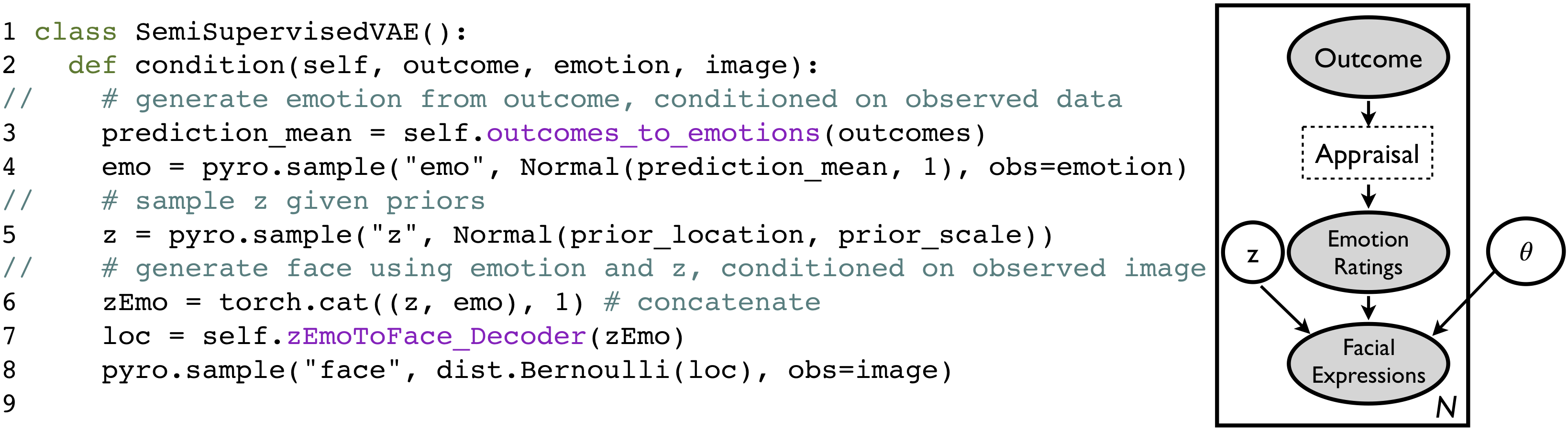}
\caption{Excerpt of Pyro code that implements a semi-supervised variant of a variational autoencoder, with graphical representation on the right. The latent variable $z$ captures aspects of the face (e.g., shape) that are emotion-irrelevant, while $\theta$ parameterizes the distribution P$_\theta$(\emph{face}$|$\emph{emotion}, $z$). Here, $\theta$ are weights in a neural network within the \texttt{Decoder()} function ($l$. 7). This code builds off Fig. \ref{fig:code1} by generating an emotion conditioned on the observed outcome and emotion ratings ($l$. 3-4), sampling $z$ from its priors ($l$. 5), and generating a face conditioned on the observed data ($l$. 6-8).}
\label{fig:code2}
\end{figure*}

Another way of modeling uncertainty, as taken in Bayesian regression, is to model the prior distributions over the model parameters. This is useful in modeling individual or group differences in appraisal. For example, people reason very differently about the emotions of someone close to them than a stranger, and we showed that we could capture this via differences in model parameters \cite{ong2018happier}. In Bayesian regression, one can hypothesize that there are several populations of people, each with their own distribution of model parameters---perhaps males tend to reason differently from females. One posits prior distributions over parameters, then sample regression models from those distributions, conditioned on the data. From the data, one can infer the population distribution of model parameters, which could provide insight into group or individual differences in emotion reasoning (We have an example in our repository).

Representing appraisal as a probabilistic program allows many exciting extensions. For example, the appraisal function can take as input an estimate of the agent's beliefs and desires. Depending on the model specification, one could first infer beliefs and desires, and then pass them into an appraisal function to reason about latent emotions \cite{demelo2014reading}, or one could jointly infer beliefs, desires and appraisals conditioned on the data \cite{wu2018rational}. We can also define richer representations for emotions, and richer computation on these representations. In the example above, the return value of the \texttt{compute\_appraisal()} function is a real-valued vector (or tensor) representing emotion intensity: Such a function could instead return an emotion ``object" that contains attributes like the \emph{target} of the emotion. To achieve this, we would have to define a space of possible targets (which could be constrained to the specific context) so that we can sample from this space during inference, and we would have to embed the target information within the appraisal representation. Although implementing this would still require effort, it seems to us that probabilistic programming offers the most plausible route to success among existing approaches.

\subsection{Learning Emotion Recognition from Faces}

Next, we consider how to easily integrate learning from high-dimensional data. For most affective computing applications, we are interested in the mapping from emotions to observable behavior like facial expressions. While there are theories mapping emotions to facial expression (e.g., the Emotion-Facial Action Coding System or EM-FACS; modified from \cite{ekman1978manual}), they may not be exhaustive, and implementation brings its own set of engineering challenges \cite{cohn2014automated}. Most researchers prefer to \emph{learn} the mapping from high-dimensional facial expression images to emotions, rather than hand-specifying this transformation.

We can extend the code above to specify a generative model from \textbf{emotions} to \textbf{facial expressions}, P$_\theta$(\emph{face}$|$\emph{emotion}), and to learn the parameters $\theta$ of such a model from data. For example, we can use a convolutional neural network to model P$_\theta$(\emph{face}$|$\emph{emotion}, $z$), where $z$ is a latent vector that captures aspects of the face that are not determined by the emotion (e.g., face shape, gender). The parameters $\theta$ can be learnt via stochastic variational inference (SVI) \cite{hoffman2013stochastic}. Modern probabilistic programming languages are able to perform SVI \emph{automatically} with a small amount of input from the modeler. SVI historically required the derivation of a quantity called the evidence lower bound (ELBO)---the ELBO is maximized during training, much like how a loss function is minimized during many machine learning approaches. In practice, the ELBO contains the posterior distribution (e.g. P($z|$\emph{face})), which is often intractable, but can be approximated with variational distributions (in our case, $q$($z|$\emph{face})) that can also be parameterized by neural networks. Trained in this manner, the model is a semi-supervised variant of the variational autoencoder (VAE) \cite{kingma2014auto}, a popular generative model that has received significant attention in the deep learning community.

Building this model in Pyro is relatively straightforward (Fig. \ref{fig:code2}), and again highlights how the complexities of inference and optimization are orthogonal to model specification. In the model, using a logic similar to Fig. \ref{fig:code1}, we sample the $z$'s from some priors (that could also be learnt). We then specify a \texttt{Decoder()} function as a neural network that takes the $z$'s and emotions and generates a corresponding face. Within PyTorch, a fairly complicated neural network can be specified in a few lines of code (see our repository).

\begin{figure*}[!bt]
\centering
\includegraphics[width=.8\textwidth]{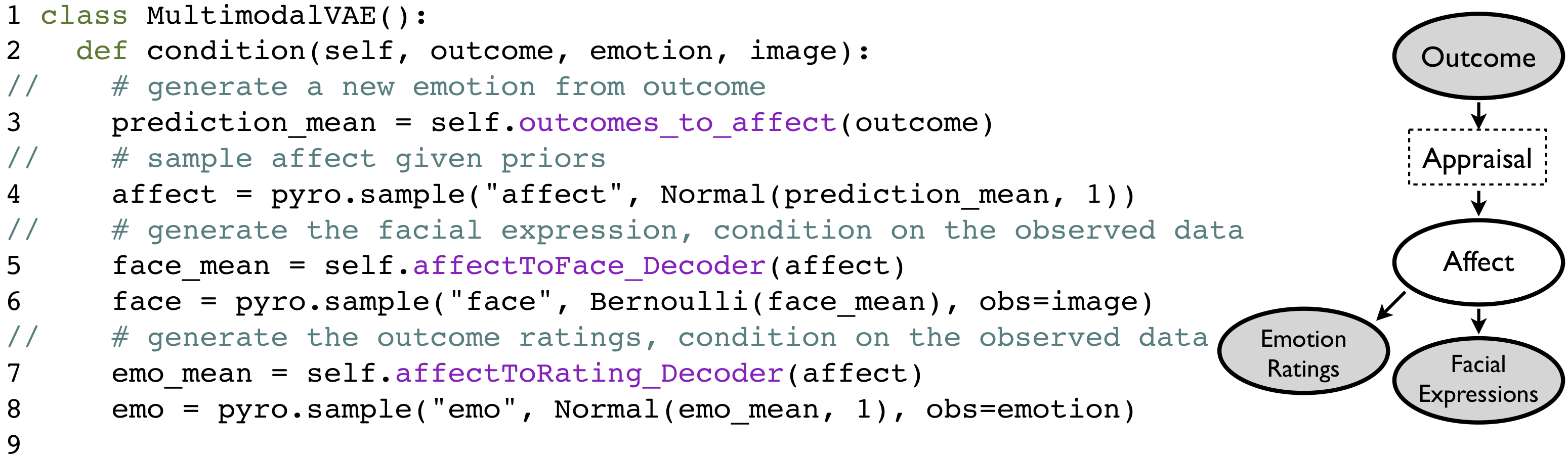}
\caption{Excerpt of Pyro code that implements a multimodal variational autoencoder. For simplicitly, we removed the plate notation and omit the variational parameters ($\theta$) that parameterize the distributions.}
\label{fig:code3}
\end{figure*}

This model can be further improved by letting the model learn the latent emotion space. Thus far, we have treated emotion as an observable variable, as we operationalized emotions using ratings that people provided, either after an agent experienced an outcome, or after an agent shows an emotional facial expression. This is usually standard practice in many supervised learning paradigms (see reviews of \cite{zeng2009survey}), where we train a model to predict an observed $y$ given variables $x$ and provide the model with fully-observed pairs of ($x$,$y$). However, in operationalizing emotion as a set of Likert scale ratings (as we have done), or a categorical label (as many classification tasks do), we limit the model to learn only the emotions that we, the modelers, chose, and only in the manner we specify. Concretely, some emotion theories have argued that there is a latent affect space that characterizes affective phenomena, and that most of the variance in emotion ratings or emotion concepts can be captured by a low-dimensional representation, such as two-dimensional space containing Valence and Arousal \cite{russell1980circumplex}, or three dimensions or Valence, Arousal, and Dominance \cite{mehrabian1996pleasure}.

To build this model, we adapt the structure of a recently-proposed Multimodal Variational Autoencoder \cite{wu2018multimodal}. In this variant of the VAE (Fig. \ref{fig:code3}), we posit a latent affect space. Outcomes, via appraisal, give rise to changes in the agent's affect, which in turn give rise to the agent's facial expressions. When participants report on the agent's affect by providing emotion ratings, they are mapping their estimate of the agent's affect onto lay emotion concepts via some mapping P(\emph{rating} $|$ \emph{affect}). Note that the rating space has dimensions of discrete emotions like \emph{happiness}, \emph{sadness} (i.e., the dimensions of the Likert scales that we asked participants to report), but the affect space is one that we allow the model to learn from the data. The model might also learn latent features about the faces, outcomes, or ratings that are unrelated to emotions, such as gender, face shape, much like the latent variable $z$ that we added into the model in Fig. \ref{fig:code2}.

The multimodal VAE model provides some flexibility in performing inference. Because of the causal dependencies in the graphical model (Fig. \ref{fig:code3}, right), each of the modalities are mutually independent given the latent affect. Thus, the model can deal effectively with any subset of the modalities. Specifically, it can do inference over emotion ratings given examples with only facial expressions, only outcomes, or examples with both facial expressions and outcomes.
This is a big advantage over many existing multimodal emotion recognition systems that cannot flexibly deal with incomplete data (i.e., observations with missing modalities).

\subsection{Using probabilistic programs as generative models}

\begin{figure}[!bt]
\centering
\includegraphics[width=.75\columnwidth]{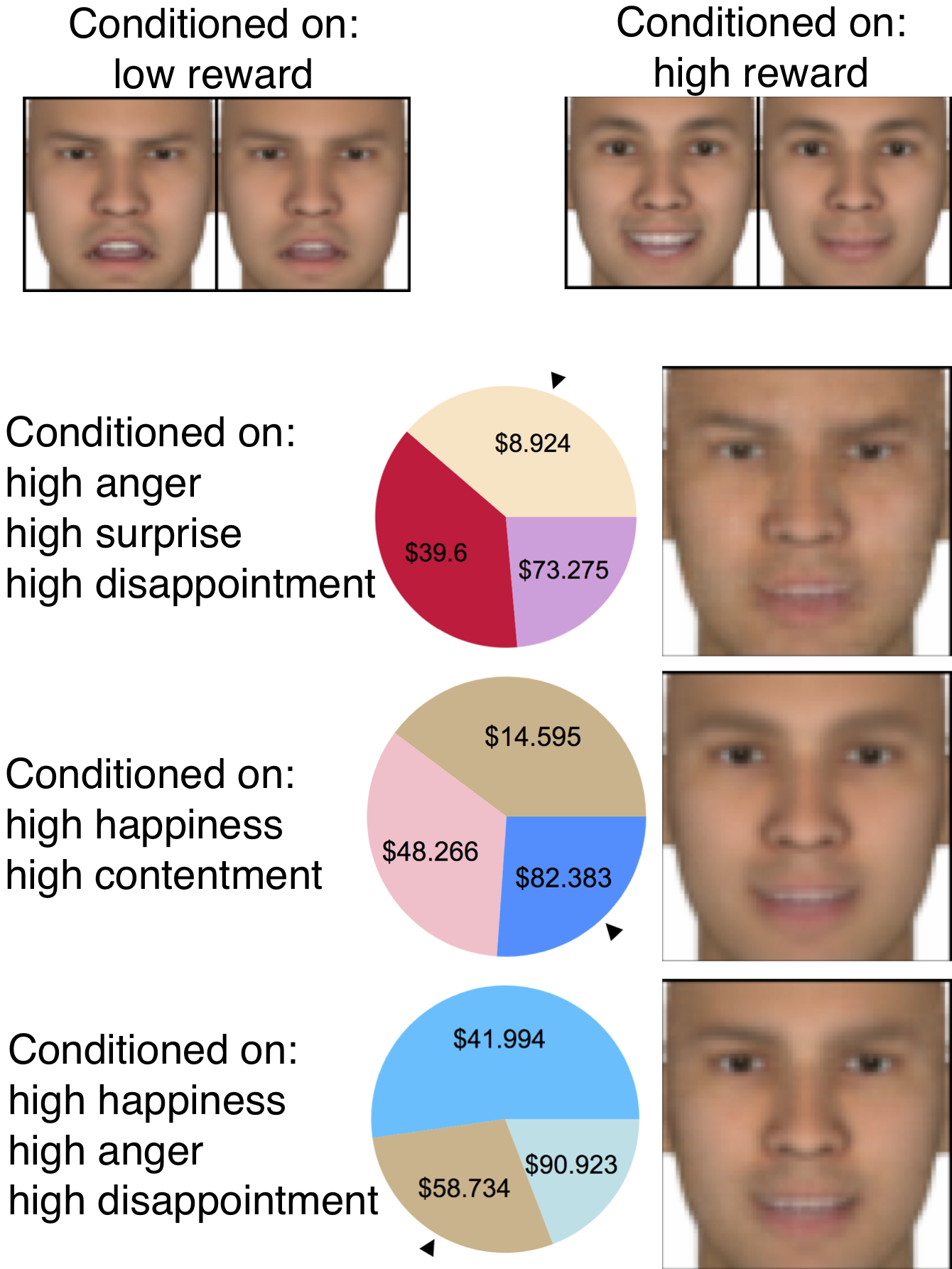}
\caption{Samples from a multimodal VAE. Top: sampled faces conditional on very low and high reward outcomes. Bottom: sampled faces and gamble outcomes, conditioned on emotion ratings. The black triangle on each wheel indicates the outcome of that gamble.}
\label{fig:ppOutput}
\end{figure}

All the models discussed here are generative models that allow us to sample new examples from the model: Indeed, sampling and conditioning is part of the inference process. In particular, if we consider the multimodal VAE model from Fig. \ref{fig:code3}, we can randomly sample a value of affect in the latent space (i.e., we can sample from the prior, often a Gaussian sphere), and generate the outcomes, faces, and emotion ratings that are associated with that value of ``affect". We can also perform conditional sampling: we can condition the model on a particular set of emotion ratings (e.g. high on happiness, high on surprise), and have the model generate the outcomes and faces that are most likely associated with those ratings. In Fig. \ref{fig:ppOutput}, we show some conditional samples from the model. Given that we only showed it a small set of faces, the model is able to reproduce the faces fairly well, and in fact can generate new faces that correspond to a novel outcome or an uncommon set of ratings.

In these examples, we have avoided discussing inference in these models for a number of reasons. First, we chose to focus on motivating the model specification from emotion theory. Second, Pyro---and PyTorch---abstracts away the optimization and inference algorithms from the affective computing modeler using high-level functions. Indeed, there exist powerful routines like stochastic variational inference \cite{hoffman2013stochastic} that solve or approximately solve the types of problems that arise in affective computing. Third, we show that the modeler can leverage state-of-the-art research done in artificial intelligence and deep learning, such as easily adapting the semi-supervised and multimodal \cite{wu2018multimodal} variants of the variational autoencoder \cite{kingma2014auto}. This can be done without necessarily getting bogged down by the details of how and why they work. For readers interested in the details of the implementation, we provide more details about inference, and links to literature and tutorials, in our code repository.

\section{General Discussion}

In this paper, we propose that probabilistic programming offers a principled and theory-driven, yet flexible and computationally efficient manner of specifying affective computing models. We draw inspiration from recent lay theories of mind and lay theories of emotion implemented using probabilistic graphical approaches, as well as two recent examples of non-affective models implemented in probabilistic programming, to propose a basic framework for modeling emotions using probabilistic programs. We provided illustrative code, written for conceptual clarity rather than predictive performance, about how one might model appraisal (reasoning about emotions from outcomes that occurred), emotion recognition from images, and inference about emotion from multiple channels. 

We hope that, by providing open-source code in a modern probabilistic programming language, we lower the barrier to entry for two groups of researchers. The first includes emotion theorists who want a standardized approach to specifying computational models of emotion, which is both easy to learn yet flexible enough to represent complex theory. For this first group, being able to leverage optimization and deep learning libraries to learn efficiently from large data is also an added bonus. The second group includes machine learning researchers and computer scientists familiar with deep learning who want to build more psychologically-grounded models, by offering a modeling paradigm that elegantly specifies scientific theory. This will hopefully provide a bridge to synergize efforts in affective computing from both data-driven and theory-driven approaches.

\subsection{Boundaries of the approach}

The probabilistic programming approach is suitable to implement a wide class of models. One way to characterize this is by borrowing David Marr's \cite{marr1982vision} classic proposal that researchers can understand computation (and cognition) at three complementary ``levels of analyses". The highest-level, the ``computational level", focuses on understanding the goal of the computation: what are the inputs needed to produce the output of a computation. For emotion recognition: what are the contextual cues and behaviour needed for inferring someone's emotions? Probabilistic approaches, as they focus on building causal models of the world, are well-suited to models at this level of analysis: Indeed, all of the models discussed in this paper are framed at this level.

By contrast, probabilistic approaches tend to be more agnostic about Marr's second, ``algorithmic" level of analysis. Models at this level are concerned with the process of transforming the inputs of the computation into the output. Questions at this level include: how do people scan facial expressions for emotional information---eyes, then mouth? How fast do people make these judgments? How long do emotion episodes last? At the present moment, most probabilistic approaches, including probabilistic programming, do not make strong commitments to process-level models, or resultant behaviour like eye-tracking and reaction times. It remains to be seen whether future probabilistic approaches can make stronger claims at this level of analysis. This is in contrast with, for example, cognitive and emotion architectures \cite{dias2014fatima, marinier2009computational, marsella2009ema}, which usually take a strong theoretical stance on the dynamics of emotion processes, such as defining fixed-interval cognitive processing cycles and similar constraints on emotional dynamics. (To finish the discussion, probabilistic approaches have even less to say about Marr's third, ``implementation" level, which is concerned with how computation is implemented in the brain or other physical systems. No one of these ``levels" is superior to the others: They answer different questions, and are all complementary to understanding computation.)

\subsection{Connection to other approaches}

In Sections 2 and 3, we discussed how the probabilistic programming approach naturally implements intuitive theories (or probabilistic graphical approaches). It is also compatible with many non-probabilistic models. For example, Ortony, Clore and Collins' \cite{ortony1990cognitive} model of appraisal (logical rules) can be easily modified and implemented in a probabilistic program. Probabilistic programs can also implement dimensional \cite{mehrabian1996pleasure, russell1980circumplex} or other \cite{yannakakis2017ordinal} representations of emotions and affect, and be used to compare competing theories. 

As mentioned in Section 5.1, probabilistic programming approaches focus on a different ``level of analysis" as process-level models and architectures that describe emotion dynamics \cite{dias2014fatima, marinier2009computational, marsella2009ema}. Indeed, we think that in the future, models built using the probabilistic programming approach will have to interface with architectural models when it comes down to defining dynamics. This should be possible given the modularity of both architectures and probabilistic programming. This might also allow us to model more task-general aspects of cognition and emotion using probabilistic programming in the future.

Probabilistic programming can also leverage deep neural network architectures. For example, Figs. \ref{fig:code2} and \ref{fig:code3} illustrate variants of a deep generative model. The compositionality of probabilistic programming allows different ``components" of the model to be represented using deep networks \cite{calvo2010affect, gunes2010automatic, jaimes2007multimodal, zeng2009survey}, but embedded within a larger theoretical model.

\subsection{Future Potential}

Probabilistic programming may help offer integration across the field of affective computing. Having a common modeling paradigm---and ideally, a common programming language---makes it easy to scientifically test competing theories. Because theories are represented as modular programs, one can easily substitute different theories (via substituting different chunks of code) within the same framework. For example, one can test different formulations of appraisal, or different representations, within this common framework. Many researchers \cite{marsella2010computational, reisenzein2013computational} have repeatedly echoed the need for a common platform to synergize research efforts across the field. While previous cognitive architectures like SOAR had seemed to promise such a common platform on which to build and test cognitive and emotion theories, the (lack of) uptake of these architectures outside the groups that developed them---relative to the uptake of machine learning and deep learning approaches to emotion---suggests that there may be numerous barriers to adoption for other scientists. Perhaps the solution lies in having accessible software packages written in popular languages supported by a vibrant development community.

Probabilistic programming may also inform greater insights into human psychology. If we train a probabilistic program to model how people recognize emotions from facial expressions, we can learn not only what facial features people tend to use in emotion judgments, but also people's relative weighting of those features, which may vary by context or individual differences. But beyond learning people's ``model parameters", probabilistic programming also allows us to refine theories. For example, researchers can specify an appraisal theory as the model's prior ``knowledge". The model can update these appraisal mappings to match empirical data, or to learn new appraisals (e.g., using tools from Bayesian non-parametrics). Researchers can then query the programs to further refine their theory. This is especially necessary today when there is too much data for researchers to specify everything necessary for modelling: We need models that can not only implement psychological theory, but also learn to \emph{add to} and refine existing theories.

As probabilistic programs are generative models, they hold promise for applications that require emotion generation, such as in virtual characters \cite{swartout2006toward}. We showed simple code that could learn to generate an emotional face conditioned on receiving a particular outcome. Obviously, this model needs a lot more data to generate realistic emotional expressions, but it could already generate emotion-appropriate faces without a programmer telling it what a smile is. This approach of learning parameters of a generative model from rater data can augment existing emotion-generation models that rely on hand-tuned expressions.

We can take the idea of emotion generation one step further, to model \emph{goal-directed emotion generation}. Earlier in the paper, we discussed the Rational Speech Act framework \cite{goodman2016pragmatic} whereby effective communication can be modeled as nested probabilistic programs. This offers a natural extension to model \emph{communicative} theories of emotion. Emotions serve a communicative role \cite{levine2018signaling, van2010interpersonal}, and people often have goals to convey their emotions to others, for example expressing appropriate negative emotion to an employee who produced sub-par work, or strategically choosing emotions to display in a negotiation \cite{dehghani2014interpersonal}. We can apply this principle to cases where an affective computing agent has to solve the goal of correctly communicating an emotion to a human user. The agent (e.g., a virtual character) builds a model $U$ of how the user would infer the agent's emotions given the agent's emotion displays. The agent can then nest that model $U$ into its own decision-making model. Based upon the agent's inferences of the user's inference of the agent's emotions, the agent can then choose its behavior to maximize the probability that the user arrives at the correct conclusion. Thus, by leveraging the compositional nature of probabilistic programs, we can embed communicative theories and goals as nested models of agents reasoning about human users reasoning about agents.

Further extensions may also allow the modelling of complex phenomena like behavioural regulation and deception. For example, people learn to regulate their emotions depending on the social context. An individual insulted at a party may not immediately act upon their anger (to confront their aggressor), and may instead choose to suppress their anger, going against the action tendencies of their current emotional state. They might even go further and fake a smile or a laugh---behaviour \emph{contrary} to their current emotional state. Current models cannot handle such complex cases, but the combination of probabilistic programming models of emotion understanding discussed here with decision-making models (e.g., POMDPs) may offer a solution. This necessitates several nested layers of reasoning made possible by compositionality: The affective computing agent reasons about how the individual feels in context, but that individual may in turn be thinking about others in their social context and their inferences about said individual.

In summary, affective computing research has made much headway over the past two decades, buoyed by the emergence of many theory-based computational models of emotion, as well as many data-driven machine learning approaches. There remains, however, much room for integration across many of these research groups and research approaches. It is our hope that the probabilistic programming paradigm, by combining the strengths of both theory-driven and data-driven approaches, may be a candidate for standardizing and unifying efforts in computational modelling of emotion and other affective phenomena.

\ifCLASSOPTIONcompsoc
  \section*{Acknowledgments}
\else
  \section*{Acknowledgment}
\fi

We thank Mike Wu and Zhi-Xuan Tan for discussion and help with coding, and Patricia Chen for feedback on the manuscript. This work was supported by the A*STAR Human-Centric Artificial Intelligence Programme (SERC SSF Project No. A1718g0048); Singapore MOE AcRF Tier 1 (No. 251RES1709) to HS; and NIH Grant 1R01MH112560-01 to JZ. This material is based on research sponsored by DARPA under agreement number FA8750-14-2-0009.

\ifCLASSOPTIONcaptionsoff
  \newpage
\fi

\bibliographystyle{IEEEtran}
\bibliography{Biblio-TAC-affCogPPL}

\begin{IEEEbiography}[{\includegraphics[width=1in,height=1.25in,clip,keepaspectratio]{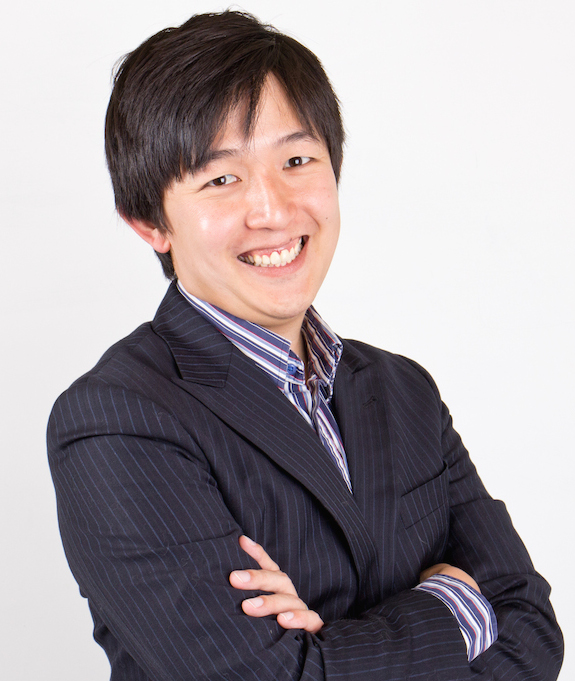}}]{Desmond C. Ong}
received his Ph.D. in Psychology and M.Sc. in Computer Science in 2017 from Stanford University. He graduated with a B.A. in Economics (\textit{summa cum laude}) and Physics (\textit{magna cum laude}), with minors in Cognitive Studies and Information Science from Cornell University in 2011. He has been a Research Scientist with the A*STAR Artificial Intelligence Initiative since 2017. His research interests include building computational models of emotion and mental state understanding, using a mix of human behavioral experiments and modeling approaches like probabilistic modeling and machine learning. He is a member of the IEEE Computer Society.
\end{IEEEbiography}

\begin{IEEEbiography}[{\includegraphics[width=1in,height=1.25in,clip,keepaspectratio]{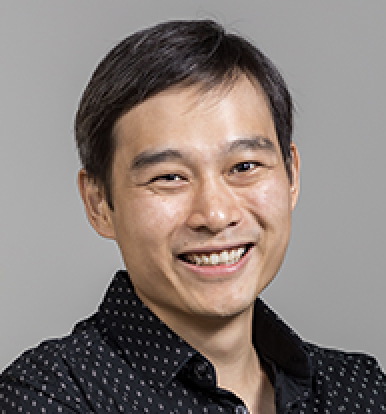}}]{Harold Soh}
received his Ph.D. in EEE from Imperial College London in 2012 where his work on assistive robotics was nominated for the Eryl Cadwaladr Davis Prize and shortlisted for the 2012 UK James Dyson Award. He did his postgraduate work under a SMART Scholar Award at MIT's SMART Center in Singapore. He is currently an Assistant Professor at the Department of Computer Science at the National University of Singapore. His primary interests are in human-robot/AI collaboration--he develops computational and machine-learning models that enable robots to better cooperate with human teammates. He is a member of the IEEE Computer Society.
\end{IEEEbiography}

\begin{IEEEbiography}[{\includegraphics[width=1in,height=1.25in,clip,keepaspectratio]{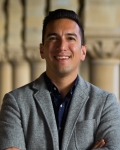}}]{Jamil Zaki}
received his Ph.D. in Psychology from Columbia University in 2010 and did his postdoctoral work at Harvard University. Since 2012, he has been an Assistant Professor of Psychology at Stanford University. He has won numerous awards, such as the 2017 Sage Young Scholar Award, a 2016 Early Career Award from the Society for Social Neuroscience, a 2015 NSF CAREER Award, and a 2015 Janet T. Spence Award for Transformative Early Career Contribution and a 2013 Rising Star award, both from the Association for Psychological Science. His research interests include empathy and emotion understanding.
\end{IEEEbiography}

\begin{IEEEbiography}[{\includegraphics[width=1in,height=1.25in,clip,keepaspectratio]{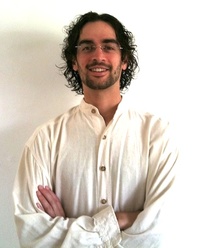}}]{Noah D. Goodman}
received his Ph.D. in Mathematics from the University of Texas at Austin, and did his postdoctoral work in the Department of Brain and Cognitive Sciences at the Massachusetts Institute of Technology. He is an Associate Professor in the Departments of Psychology and Computer Science at Stanford University, and also holds a courtesy appointment in the Department of Linguistics. His work has been recognized by several awards such as the Alfred P Sloan Research Fellow in Neuroscience and a J. S. McDonnell Foundation Scholar. His research interests include language understanding, social reasoning, and concept learning. In addition, he builds enabling technologies for probabilistic modeling, such as probabilistic programming languages.
\end{IEEEbiography}





\end{document}